\documentclass[11pt]{article} 
\usepackage{rldmsubmit,palatino}
\usepackage{graphicx} 
\usepackage{enumerate, amsmath, amsthm, amssymb, algorithm, algorithmic, pifont, fullpage, csquotes, dashrule, tikz, bbm, booktabs, bm, verbatim, url, dsfont}
\usepackage[framemethod=TikZ]{mdframed}
\usepackage[numbers]{natbib}
\usepackage[colorlinks]{hyperref}
\hypersetup{
     colorlinks = true,
     linkcolor = blue,
     anchorcolor = blue,
     citecolor = red,
     filecolor = blue,
     urlcolor = blue
     }
\usepackage{caption}
\usepackage{subcaption}

\definecolor{dblue}{RGB}{98, 140, 190}
\definecolor{dlblue}{RGB}{216, 235, 255}
\definecolor{dgreen}{RGB}{124, 155, 127}
\definecolor{dpink}{RGB}{207, 166, 208}
\definecolor{dyellow}{RGB}{255, 248, 199}
\definecolor{dgray}{RGB}{46, 49, 49}

\newcommand{\durl}[1]{\textcolor{dblue}{\underline{\url{#1}}}}

\newcommand{\mc}[1]{\mathcal{#1}}

\newcommand{\bE}{\mathbb{E}}

\newcommand{\bR}{\mathbb{R}}
\newcommand{\bP}{\mathbb{P}}

\newcommand{\bI}{\mathbb{I}}
\newcommand{\bH}{\mathbb{H}}
\newcommand{\bN}{\mathbb{N}}

\newcommand{\kl}[2]{D_{\mathrm{KL}}(#1\text{ }||\text{ }#2)}

\newcommand{\ra}{\rightarrow}

\newmdenv[
  topline=false,
  bottomline=false,
  rightline = false,
  leftmargin=10pt,
  rightmargin=0pt,
  innertopmargin=0pt,
  innerbottommargin=0pt
]{innerproof}

\newcounter{DaveDefCounter}
\newtheorem{example}{Example}

\newif\ifsubmit
\submitfalse
\ifsubmit
\newcommand{\dnote}[1]{}
\newcommand{\bnote}[1]{}
\else
\newcommand{\dnote}[1]{\textcolor{blue}{Dilip: #1}}
\newcommand{\bnote}[1]{\textcolor{orange}{Ben: #1}}
\fi

\title{Between Rate-Distortion Theory \& Value Equivalence in Model-Based Reinforcement Learning}

\author{
Dilip Arumugam \\
Department of Computer Science\\
Stanford University\\
\texttt{dilip@cs.stanford.edu} \\
\And
Benjamin Van Roy \\
Department of Electrical Engineering\\
Department of Management Science \& Engineering\\
Stanford University\\
\texttt{bvr@stanford.edu} \\
}

\begin{document}

\maketitle

\begin{abstract}
The quintessential model-based reinforcement-learning agent iteratively refines its estimates or prior beliefs about the true underlying model of the environment. Recent empirical successes in model-based reinforcement learning with function approximation, however, eschew the true model in favor of a surrogate that, while ignoring various facets of the environment, still facilitates effective planning over behaviors. Recently formalized as the value equivalence principle, this algorithmic technique is perhaps unavoidable as real-world reinforcement learning demands consideration of a simple, computationally-bounded agent interacting with an overwhelmingly complex environment. In this work, we entertain an extreme scenario wherein some combination of immense environment complexity and limited agent capacity entirely precludes identifying an exactly value-equivalent model. In light of this, we embrace a notion of approximate value equivalence and introduce an algorithm for incrementally synthesizing \textit{simple} and \textit{useful} approximations of the environment from which an agent might still recover near-optimal behavior. Crucially, we recognize the information-theoretic nature of this lossy environment compression problem and use the appropriate tools of rate-distortion theory to make mathematically precise how value equivalence can lend tractability to otherwise intractable sequential decision-making problems. 
\end{abstract}


\acknowledgements{The authors gratefully acknowledge Christopher Grimm for initial discussions that provided an impetus for this work. Financial support from Army Research Office (ARO) grant W911NF2010055 is gratefully acknowledged.}

\startmain 

\section{Problem Formulation}

We formulate a sequential decision-making problem as an episodic, finite-horizon Markov Decision Process (MDP)~\citep{bellman1957markovian,Puterman94} defined by $\mc{M} = \langle \mc{S}, \mc{A}, \mc{R}, \mc{T}, \beta, H \rangle$. $\mc{S}$ denotes a set of states, $\mc{A}$ is a set of actions, $\mc{R}:\mc{S} \times \mc{A} \ra [0,1]$ is a deterministic reward function providing evaluative feedback signals (in the unit interval) to the agent, $\mc{T}:\mc{S} \times \mc{A} \ra \Delta(\mc{S})$ is a transition function prescribing distributions over next states, $\beta \in \Delta(\mc{S})$ is an initial state distribution, and $H \in \bN$ is the maximum episode length or horizon. 

Let $(\Omega, \mc{F}, \bP)$ be a probability space. As is standard in Bayesian reinforcement learning, both the transition function and reward function are not known to the agent and are consequently treated as random variables. With all other MDP components known a priori, the randomness in the model fully accounts for the randomness in the MDP, which is also a random variable. We denote by $\mc{M}^\star$ the true MDP with model $(\mc{R}^\star, \mc{T}^\star)$ that the agent interacts with and attempts to solve over the course of $K$ episodes. Within each episode, the agent acts for exactly $H$ steps beginning with an initial state $s_1 \sim \beta$. For each $h \in [H]$, the agent observes the current state $s_h \in \mc{S}$, selects action $a_h \sim \pi_h(\cdot \mid s_h) \in \mc{A}$, enjoys a reward $r_h = \mc{R}(s_h,a_h) \in [0,1]$, and transitions to the next state $s_{h+1} \sim \mc{T}(\cdot \mid s_h, a_h) \in \mc{S}$.

A stationary, stochastic policy for timestep $h \in [H]$, $\pi_h:\mc{S} \ra \Delta(\mc{A})$, encodes a pattern of behavior mapping individual states to distributions over possible actions. Letting $\{\mc{S} \ra \Delta(\mc{A})\}$ denote the class of all stationary, stochastic policies, a non-stationary policy $\pi = (\pi_1,\ldots,\pi_H) \in \{\mc{S} \ra \Delta(\mc{A})\}^H$ is a collection of exactly $H$ stationary, stochastic policies  whose overall performance in any MDP $\mc{M}$ at timestep $h \in [H]$ when starting at state $s \in \mc{S}$ and taking action $a \in \mc{A}$ is assessed by its associated action-value function $Q^\pi_{\mc{M},h}(s,a) = \bE\left[\sum\limits_{h'=h}^H \mc{R}(s_{h'},a_{h'}) \bigm| s_h = s, a_h = a\right]$, where the expectation integrates over randomness in the action selections and transition dynamics. Taking the value function as $V^\pi_{\mc{M},h}(s) = \bE_{a \sim \pi_h(\cdot \mid s)}\left[Q^\pi_{\mc{M},h}(s,a)\right]$, we define the optimal policy $\pi^\star = (\pi^\star_1,\pi^\star_2,\ldots,\pi^\star_H)$ as achieving supremal value $V^\star_{\mc{M},h}(s) = \sup\limits_{\pi \in \{\mc{S} \ra \Delta(\mc{A})\}^H} V^\pi_{\mc{M},h}(s)$ for all $s \in \mc{S}$, $h \in [H]$. We let $\tau_k = (s^{(k)}_1, a^{(k)}_1, r^{(k)}_1, \ldots,s^{(k)}_{H}, a^{(k)}_{H}, r^{(k)}_{H}, s^{(k)}_{H+1})$ be a random variable denoting the trajectory experienced by the agent in the $k$th episode. Meanwhile, $H_k = \{\tau_1,\tau_2,\ldots, \tau_{k-1}\} \in \mc{H}_k$ is a random variable representing the entire history of the agent's interaction within the environment at the start of the $k$th episode. Abstractly, a reinforcement-learning algorithm is a sequence of non-stationary policies $(\pi^{(1)},\ldots,\pi^{(K)})$ where, for each episode $k \in [K]$, $\pi^{(k)}:\mc{H}_k \ra \{\mc{S} \ra \Delta(\mc{A})\}$ is a function of the current history $H_k$. We note that no further restrictions on the state-action space $\mc{S} \times \mc{A}$, such as finiteness, have been made; notably, through our use of information theory, our algorithm may operate on any finite-horizon, episodic MDP although we leave the question of how to practically instantiate our algorithm for concrete settings of interest 
to future work. 

\section{Rate-Distortion Theory}

We here provide a brief, high-level overview of rate-distortion theory~\citep{shannon1959coding} and encourage readers to consult \citep{cover2012elements} for more details. A lossy compression problem consumes as input a fixed information source $\bP(X \in \cdot)$ and a distortion function $d: \mc{X} \times \mc{Z} \ra \bR_{\geq 0}$ which quantifies the loss of fidelity by using a compression $Z$ in place of the original $X$. Then, for any distortion threshold $D \in \bR_{\geq 0}$, the rate-distortion function quantifies the fundamental limit of lossy compression as $$\mc{R}(D) = \inf\limits_{Z \in \Lambda} \bI(X;Z) \triangleq \inf\limits_{Z \in \Lambda} \bE\left[\kl{\bP\left(X \in \cdot \mid Z\right)}{\bP(X \in \cdot)}\right] \qquad \Lambda \triangleq \{Z: \Omega \ra \mc{Z} \mid \bE\left[d(X,Z)\right] \leq D\},$$ where $\bI(X;Z)$ denotes the mutual information and the infimum is taken over all random variables $Z$ that incur bounded expected distortion, $\bE\left[d(X,Z)\right] \leq D$. Naturally, $\mc{R}(D)$ represents the minimum number of bits of information that must be retained from $X$ in order to achieve this bounded expected loss of fidelity. In keeping with the previous problem formulation, which does not assume discrete random variables, we note that the rate-distortion function is well-defined for information source and channel output random variables taking values on abstract alphabets~\citep{csiszar1974extremum}. Moreover, the problem of computing the rate-distortion function along with the channel that achieves its infimum is well-studied and solved by the classic Blahut-Arimoto algorithm~\citep{blahut1972computation,arimoto1972algorithm}, which is computationally feasible for discrete channel outputs. 

Just as in past work that studies satisficing in multi-armed bandit problems~\citep{russo2022satisficing,arumugam2021deciding,arumugam2021the}, we use rate-distortion theory to formalize and identify a simplified MDP $\widetilde{\mc{M}}_k$ that the agent will attempt to learn over the course of each episode $k \in [K]$. The episode dependence arises from utilizing the agent's current beliefs over the true MDP $\bP(\mc{M}^\star \in \cdot \mid H_k)$ as an information source to be lossily compressed.

\section{The Value Equivalence Principle}

As outlined in the previous section, the second input for a well-specified lossy-compression problem is a distortion function prescribing non-negative real values to realizations of the information source and channel output random variables $(\mc{M}^\star, \widetilde{\mc{M}})$ that quantify the loss of fidelity incurred by using $\widetilde{\mc{M}}$ in lieu of $\mc{M}^\star$. To define this function, we will leverage an approximate notion of value equivalence~\citep{grimm2020value,grimm2021proper}. For any arbitrary MDP $\mc{M}$ with model $(\mc{R},\mc{T})$ and any stationary, stochastic policy $\pi:\mc{S} \ra \Delta(\mc{A})$, define the Bellman operator $\mc{B}^\pi_\mc{M}: \{\mc{S} \ra \bR\} \ra \{\mc{S} \ra \bR\}$ as follows: $\mc{B}^\pi_\mc{M}V(s) \triangleq \bE_{a \sim \pi(\cdot \mid s)}\left[\mc{R}(s,a) + \bE_{s' \sim \mc{T}(\cdot \mid s, a)}\left[ V(s')\right]\right].$ The Bellman operator is a foundational tool in dynamic-programming approaches to reinforcement learning~\citep{bertsekas1995dynamic} and gives rise to the classic Bellman equation: for any MDP $\mc{M} = \langle \mc{S}, \mc{A}, \mc{R}, \mc{T}, \beta, H \rangle$ and any non-stationary policy $\pi = (\pi_1,\ldots,\pi_H)$, the value functions induced by $\pi$ satisfy $V^\pi_{\mc{M},h}(s) = \mc{B}^{\pi_h}_{\mc{M}}V^\pi_{\mc{M},h+1}(s),$ for all $h \in [H]$ and with $V^\pi_{\mc{M},H+1}(s) = 0$, $\forall s \in \mc{S}$. 

For any two MDPs $\mc{M} = \langle \mc{S}, \mc{A}, \mc{R}, \mc{T}, \beta, H \rangle$ and $\widehat{\mc{M}} = \langle \mc{S}, \mc{A}, \widehat{\mc{R}}, \widehat{\mc{T}}, \beta, H \rangle$, \citet{grimm2020value} define a notion of equivalence between them despite their differing models. For any policy class $\Pi \subseteq \{\mc{S} \ra \Delta(\mc{A})\}$ and value function class $\mc{V} \subseteq \{\mc{S} \ra \bR\}$, $\mc{M}$ and $\widehat{\mc{M}}$ are value equivalent with respect to $\Pi$ and $\mc{V}$ if and only if $\mc{B}^\pi_{\mc{M}}V = \mc{B}^\pi_{\widehat{\mc{M}}}V$, $\forall \pi \in \Pi, V \in \mc{V}.$ In words, two different models are deemed value equivalent if they induce identical Bellman updates under any pair of policy and value function from $\Pi \times \mc{V}$. \citet{grimm2020value} prove that when $\Pi = \{\mc{S} \ra \Delta(\mc{A})\}$ and $\mc{V} = \{\mc{S} \ra \bR\}$, the set of all exactly value-equivalent models is a singleton set containing only the true model of the environment. The key insight behind value equivalence, however, is that practical model-based reinforcement-learning algorithms need not be concerned with modeling every granular detail of the underlying environment and may, in fact, stand to benefit by optimizing an alternative criterion besides the traditional maximum-likelihood objective~\citep{silver2017predictron,oh2017value,schrittwieser2020mastering}. Indeed, by restricting focus to decreasing subsets of policies $\Pi \subset \{\mc{S} \ra \Delta(\mc{A})\}$ and value functions $\mc{V} \subset \{\mc{S} \ra \bR\}$, the space of exactly value-equivalent models is monotonically increasing. 

For brevity, let $\mathfrak{R} \triangleq \{\mc{S} \times \mc{A} \ra [0,1]\}$ and $\mathfrak{T} \triangleq \{\mc{S} \times \mc{A} \ra \Delta(\mc{S})\}$ denote the classes of all reward functions and transition functions, respectively. Recall that, with all uncertainty in $\mc{M}^\star$ entirely driven by its model, we may think of the support of $\mc{M}^\star$ as $\mathfrak{M} \triangleq \mathfrak{R} \times \mathfrak{T}$. We define a distortion function on pairs of MDPs $d:\mathfrak{M} \times \mathfrak{M} \ra \bR_{\geq 0}$ for any $\Pi \subseteq \{\mc{S} \ra \Delta(\mc{A})\}$, $\mc{V} \subseteq \{\mc{S} \ra \bR\}$ as $$d_{\Pi,\mc{V}}(\mc{M},\widehat{\mc{M}}) = \sup\limits_{\substack{\pi \in \Pi \\ V \in \mc{V}}} ||\mc{B}^\pi_{\mc{M}}V - \mc{B}^\pi_{\widehat{\mc{M}}}V||_\infty^2 = \sup\limits_{\substack{\pi \in \Pi \\ V \in \mc{V}}} \left(\max\limits_{s \in \mc{S}} |\mc{B}^\pi_{\mc{M}}V(s) - \mc{B}^\pi_{\widehat{\mc{M}}}V(s)| \right)^2.$$ In words, $d_{\Pi,\mc{V}}$ is the supremal squared Bellman error between MDPs $\mc{M}$ and $\widehat{\mc{M}}$ across all states $s \in \mc{S}$ with respect to the policy class $\Pi$ and value function class $\mc{V}$.

\section{Value-Equivalent Sampling for Reinforcement Learning}

By virtue of the previous two sections, we are now in a position to define the lossy compression problem that characterizes a MDP $\widetilde{\mc{M}}_k$ that the agent will endeavor to learn in each episode $k \in [K]$ instead of the true MDP $\mc{M}^\star$. For any $\Pi \subseteq \{\mc{S} \ra \Delta(\mc{A})\}$; $\mc{V} \subseteq \{\mc{S} \ra \bR\}$; $k \in [K]$; and $D \geq 0$, we define the rate-distortion function
\begin{align}
    \mc{R}^{\Pi,\mc{V}}_k(D) = \inf\limits_{\widetilde{\mc{M}} \in \Lambda} \bI_k(\mc{M}^\star; \widetilde{\mc{M}}) \triangleq \inf\limits_{\widetilde{\mc{M}} \in \Lambda} \bE\left[\kl{\bP(\mc{M}^\star \in \cdot \mid \widetilde{\mc{M}} , H_k)}{\bP(\mc{M}^\star \in \cdot \mid H_k)} \mid H_k\right],
    \label{eq:rdf}
\end{align}
where $\Lambda \triangleq \big\{\widetilde{\mc{M}}: \Omega \ra \mathfrak{M} \mid \bE[d_{\Pi,\mc{V}}(\mc{M}^\star,\widetilde{\mc{M}}) \mid H_k] \leq D\big\}$. This rate-distortion function characterizes the fundamental limit of lossy MDP compression under our chosen distortion measure resulting in a channel that retains the minimum amount of information from the true MDP $\mc{M}^\star$ while yielding an approximately value-equivalent MDP in expectation. Observe that this distortion constraint is a notion of approximate value equivalence which collapses to the exact value equivalence of \citet{grimm2020value} as $D \ra 0$. Meanwhile, as $D \ra \infty$, we accommodate a more aggressive compression of the true MDP $\mc{M}^\star$ resulting in less faithful Bellman updates.
\vspace{-15pt}
\begin{center}
\begin{minipage}{0.41\textwidth}
\vspace{-48pt}
\begin{algorithm}[H]
   \caption{Posterior Sampling for Reinforcement Learning (PSRL)~\citep{strens2000bayesian}}
   \label{alg:psrl}
\begin{algorithmic}
   \STATE {\bfseries Input:} Prior $\bP(\mc{M}^\star \in \cdot \mid H_1)$
   \FOR{$k \in [K]$}
   \STATE Sample $M_k \sim \bP(\mc{M}^\star \in \cdot \mid H_k)$
   \STATE Get optimal policy $\pi^{(k)} = \pi^\star_{M_k}$
   \STATE Execute $\pi^{(k)}$ and get trajectory $\tau_k$
   \STATE Update history $H_{k+1} = H_k \cup \tau_k$
   \STATE Induce posterior $\bP(\mc{M}^\star \in \cdot \mid H_{k+1})$
   \ENDFOR
\end{algorithmic}
\end{algorithm}
\end{minipage}
\hfill
\begin{minipage}{0.58\textwidth}
\begin{algorithm}[H]
   \caption{Value-equivalent Sampling for Reinforcement Learning (VSRL)}
   \label{alg:vsrl}
\begin{algorithmic}
   \STATE {\bfseries Input:} Prior distribution $\bP(\mc{M}^\star \in \cdot \mid H_1)$, Distortion threshold $D \in \bR_{\geq 0}$, Distortion function $d_{\Pi,\mc{V}}: \mathfrak{M} \times \mathfrak{M} \ra \bR_{\geq 0}$
   \FOR{$k \in [K]$}
   \STATE Compute channel $\bP(\widetilde{\mc{M}}_k \in \cdot \mid \mc{M}^\star)$ achieving $\mc{R}^{\Pi,\mc{V}}_k(D)$ limit (Equation \ref{eq:rdf})
   \STATE Sample MDP $M^\star \sim \bP(\mc{M}^\star \in \cdot \mid H_k)$
   \STATE Sample compressed MDP $M_k \sim \bP(\widetilde{\mc{M}}_k \in \cdot \mid \mc{M}^\star = M^\star)$
   \STATE Compute optimal policy $\pi^{(k)} = \pi^\star_{M_k}$
   \STATE Execute $\pi^{(k)}$ and observe trajectory $\tau_k$
   \STATE Update history $H_{k+1} = H_k \cup \tau_k$
   \STATE Induce posterior $\bP(\mc{M}^\star \in \cdot \mid H_{k+1})$
   \ENDFOR
\end{algorithmic}
\end{algorithm}
\end{minipage}
\end{center}

A standard algorithm for our problem setting is widely known as Posterior Sampling for Reinforcement Learning (PSRL)~\citep{strens2000bayesian,osband2017posterior}, which we present as Algorithm \ref{alg:psrl}, while our Value-equivalent Sampling for Reinforcement Learning (VSRL) is given as Algorithm \ref{alg:vsrl}.  The key distinction between them is that, at each episode $k \in [K]$, the latter takes the posterior sample $M^\star \sim \bP(\mc{M}^\star \in \cdot \mid H_k)$ and passes it through the channel that achieves the rate-distortion limit (Equation \ref{eq:rdf}) at this episode to get the $M_k$ whose optimal policy is executed in the environment.
\vspace{-8pt}
\section{Discussion}
\vspace{-8pt}
\begin{example}[A Multi-Resolution MDP]
For a large but finite $N \in \bN$, consider a sequence of MDPs, $\{\mc{M}_n\}_{n \in [N]}$, which all share a common action space $\mc{A}$ but vary in state space ($\mc{S}_n$), reward function, and transition function. Moreover, for each $n \in [N]$, the rewards of the $n$th MDP are bounded in the interval $[0,\frac{1}{n}]$. An agent is confronted with the resulting product MDP, $\mc{M}$, defined on the state space $\mc{S}_1 \times \ldots \times \mc{S}_N$ with action space $\mc{A}$ and rewards summed across the $N$ constituent reward functions. The transition function is defined such that each action $a \in \mc{A}$ is executed across all $N$ MDPs simultaneously and the resulting individual transitions are composed to make a transition of $\mc{M}$. For any value of $N$, PSRL will persistently act to identify the transition and reward structure of all $\{\mc{M}_n\}_{n \in [N]}$.
\label{example:multi_res_mdps}
\end{example}
Example \ref{example:multi_res_mdps} presents a scenario where, as $N \uparrow \infty$, a complex environment retains a wealth of information, and yet, only a subset of that information may be within the agent's reach or even necessary for producing reasonably competent behavior. VSRL implicitly identifies a $M \ll N$ such that learning the subsequence of MDPs $\{\mc{M}_n\}_{n \in [M]}$ is sufficient for achieving a desired degree of sub-optimality.


The core impetus for this work is to recognize that, for complex environments, pursuit of the exact MDP $\mc{M}^\star$ may be an entirely infeasible goal. Consider a MDP that represents control of a real-world, physical system; learning a transition function of the associated environment, at some level, demands that the agent internalize laws of physics and motion to a reasonable degree of accuracy. More formally, take the random variable $M_1 \sim \bP(\mc{M}^\star \in \cdot \mid H_1)$ reflecting the agent's prior beliefs over $\mc{M}^\star$. Denoting $\bH(\cdot)$ as the entropy of a random variable, observe that identifying $\mc{M}^\star$ requires that a PSRL agent obtain exactly $\bH(M_1)$ bits of information from the environment which, under an uninformative prior, may either be prohibitively large and exceed the agent's capacity constraints or simply be impractical under time and resource constraints. 

\section{Conclusion}

In this work, we embrace the idea of \textit{satisficing}~\citep{russo2022satisficing,arumugam2021deciding,arumugam2021the}; as succinctly stated by Herbert A. Simon during his 1978 Nobel Memorial Lecture, ``decision makers can satisfice either by finding optimum solutions for a simplified world, or by finding satisfactory solutions for a more realistic world.'' Rather than spend an inordinate amount of time trying to recover an optimum solution to the true environment, VSRL pursues optimum solutions for a sequence of simplified environments. Future work will develop a complementary regret analysis that demonstrates how finding such optimum solutions for simplified worlds ultimately acts as a mechanism for achieving a satisfactory solution for the realistic, complex world. Naturally, the loss of fidelity between the simplified and true environments translates into a fixed amount of regret that an agent designer consciously and willingly accepts for two reasons: (1) they expect a reduction in the amount of time, data, and bits of information needed to identify the simplified environment and (2) in tasks where the environment encodes irrelevant information and exact knowledge isn't needed to achieve optimal behavior~\citep{farahmand2017value,grimm2020value,grimm2021proper}, a VSRL agent may still identify the optimal policy while maintaining greater sample efficiency than traditional PSRL.

\bibliographystyle{plainnat}
\bibliography{rldm_references}

\begin{thebibliography}{19}
\providecommand{\natexlab}[1]{#1}
\providecommand{\url}[1]{\texttt{#1}}
\expandafter\ifx\csname urlstyle\endcsname\relax
  \providecommand{\doi}[1]{doi: #1}\else
  \providecommand{\doi}{doi: \begingroup \urlstyle{rm}\Url}\fi

\bibitem[Arimoto(1972)]{arimoto1972algorithm}
Suguru Arimoto.
\newblock An algorithm for computing the capacity of arbitrary discrete
  memoryless channels.
\newblock \emph{IEEE Transactions on Information Theory}, 18\penalty0
  (1):\penalty0 14--20, 1972.

\bibitem[Arumugam and Van~Roy(2021{\natexlab{a}})]{arumugam2021deciding}
Dilip Arumugam and Benjamin Van~Roy.
\newblock Deciding what to learn: {A} rate-distortion approach.
\newblock In \emph{International Conference on Machine Learning}, pages
  373--382. PMLR, 2021{\natexlab{a}}.

\bibitem[Arumugam and Van~Roy(2021{\natexlab{b}})]{arumugam2021the}
Dilip Arumugam and Benjamin Van~Roy.
\newblock The value of information when deciding what to learn.
\newblock \emph{Advances in Neural Information Processing Systems}, 34,
  2021{\natexlab{b}}.

\bibitem[Bellman(1957)]{bellman1957markovian}
Richard Bellman.
\newblock A {M}arkovian decision process.
\newblock \emph{Journal of Mathematics and Mechanics}, pages 679--684, 1957.

\bibitem[Bertsekas(1995)]{bertsekas1995dynamic}
Dimitri~P. Bertsekas.
\newblock \emph{Dynamic Programming and Optimal Control}.
\newblock Athena Scientific, 1995.

\bibitem[Blahut(1972)]{blahut1972computation}
Richard Blahut.
\newblock Computation of channel capacity and rate-distortion functions.
\newblock \emph{IEEE Transactions on Information Theory}, 18\penalty0
  (4):\penalty0 460--473, 1972.

\bibitem[Cover and Thomas(2012)]{cover2012elements}
Thomas~M Cover and Joy~A Thomas.
\newblock \emph{Elements of {I}nformation {T}heory}.
\newblock John Wiley \& Sons, 2012.

\bibitem[Csisz{\'a}r(1974)]{csiszar1974extremum}
Imre Csisz{\'a}r.
\newblock On an extremum problem of information theory.
\newblock \emph{Studia Scientiarum Mathematicarum Hungarica}, 9, 1974.

\bibitem[Farahmand et~al.(2017)Farahmand, Barreto, and
  Nikovski]{farahmand2017value}
Amir-massoud Farahmand, Andre Barreto, and Daniel Nikovski.
\newblock Value-aware loss function for model-based reinforcement learning.
\newblock In \emph{Artificial Intelligence and Statistics}, pages 1486--1494.
  PMLR, 2017.

\bibitem[Grimm et~al.(2020)Grimm, Barreto, Singh, and Silver]{grimm2020value}
Christopher Grimm, Andre Barreto, Satinder Singh, and David Silver.
\newblock The value equivalence principle for model-based reinforcement
  learning.
\newblock \emph{Advances in Neural Information Processing Systems}, 33, 2020.

\bibitem[Grimm et~al.(2021)Grimm, Barreto, Farquhar, Silver, and
  Singh]{grimm2021proper}
Christopher Grimm, Andre Barreto, Gregory Farquhar, David Silver, and Satinder
  Singh.
\newblock Proper value equivalence.
\newblock In \emph{Thirty-Fifth Conference on Neural Information Processing
  Systems}, 2021.

\bibitem[Oh et~al.(2017)Oh, Singh, and Lee]{oh2017value}
Junhyuk Oh, Satinder Singh, and Honglak Lee.
\newblock Value prediction network.
\newblock In \emph{Proceedings of the 31st International Conference on Neural
  Information Processing Systems}, pages 6120--6130, 2017.

\bibitem[Osband and Van~Roy(2017)]{osband2017posterior}
Ian Osband and Benjamin Van~Roy.
\newblock Why is posterior sampling better than optimism for reinforcement
  learning?
\newblock In \emph{International Conference on Machine Learning}, pages
  2701--2710. PMLR, 2017.

\bibitem[Puterman(1994)]{Puterman94}
Martin~L. Puterman.
\newblock \emph{{M}arkov Decision Processes---Discrete Stochastic Dynamic
  Programming}.
\newblock John Wiley \& Sons, Inc., New York, NY, 1994.

\bibitem[Russo and Van~Roy(2022)]{russo2022satisficing}
Daniel Russo and Benjamin Van~Roy.
\newblock Satisficing in time-sensitive bandit learning.
\newblock \emph{Mathematics of Operations Research}, 2022.

\bibitem[Schrittwieser et~al.(2020)Schrittwieser, Antonoglou, Hubert, Simonyan,
  et~al.]{schrittwieser2020mastering}
Julian Schrittwieser, Ioannis Antonoglou, Thomas Hubert, Karen Simonyan, et~al.
\newblock Mastering {A}tari, {G}o, {C}hess and {S}hogi by planning with a
  learned model.
\newblock \emph{Nature}, 588\penalty0 (7839):\penalty0 604--609, 2020.

\bibitem[Shannon(1959)]{shannon1959coding}
Claude~E. Shannon.
\newblock Coding theorems for a discrete source with a fidelity criterion.
\newblock \emph{IRE Nat. Conv. Rec., March 1959}, 4:\penalty0 142--163, 1959.

\bibitem[Silver et~al.(2017)Silver, Hasselt, Hessel,
  et~al.]{silver2017predictron}
David Silver, Hado Hasselt, Matteo Hessel, et~al.
\newblock The {P}redictron: End-to-end learning and planning.
\newblock In \emph{International Conference on Machine Learning}, pages
  3191--3199. PMLR, 2017.

\bibitem[Strens(2000)]{strens2000bayesian}
Malcolm~JA Strens.
\newblock A {B}ayesian framework for reinforcement learning.
\newblock In \emph{Proceedings of the Seventeenth International Conference on
  Machine Learning}, pages 943--950, 2000.

\end{thebibliography}

\end{document}